%% file: main.tex
\documentclass{article}
\usepackage{spconf,amsmath,graphicx,amsfonts}
\usepackage{subcaption}
\usepackage{multirow}
\usepackage[linesnumbered]{algorithm2e}
\usepackage{algcompatible}
\usepackage{hyperref}
\usepackage{makecell}
\usepackage{comment}

\title{SaDI: A Self-adaptive Decomposed Interpretable Framework for Electric Load Forecasting under Extreme Events}

\name{
\begin{tabular}{c}
Hengbo Liu$^{\rm{1} \dagger}$ $\qquad$ Ziqing Ma$^{\rm{1} \dagger}$ $\qquad$ Linxiao Yang$^{\rm{1}}$ $\qquad$ Tian Zhou$^{\rm{1}}$ $\qquad$ Rui Xia$^{\rm{1}}$ \\
$\qquad$ Yi Wang$^{\rm{3}}$ $\qquad$ Qingsong Wen$^{\rm{2}}$ $\qquad$ Liang Sun$^{\rm{2} \star}$
\end{tabular}
}

\address{$^{\rm{1}}$ DAMO Academy, Alibaba Group, Hangzhou, China \\
$^{\rm{2}}$ DAMO Academy, Alibaba Group, Bellevue, US \\
$^{\rm{3}}$ The University of Hong Kong, Hong Kong, China }

\begin{document}

\maketitle



\input{sections/0_abstract}

\let\thefootnote\relax\footnotetext{$\dagger$ Equal contribution}
\let\thefootnote\relax\footnotetext{$*$ Corresponding authors}

\input{sections/1_introduction}

\input{sections/1.9_statement_of_problem}
\input{sections/2_methodology}
\input{sections/3_experiments}

\input{sections/4_conclusions}

\bibliographystyle{IEEEbib}
\bibliography{main}
\input{appendix/1_appendix}

\end{document}

%% file: sections/0_abstract.tex
\begin{abstract}
Accurate prediction of electric load is crucial in power grid planning and management. In this paper, we solve the electric load forecasting problem under extreme events such as scorching heats. One challenge for accurate forecasting is the lack of training samples under extreme conditions. Also load usually  changes dramatically in these extreme conditions, which calls for interpretable model to make better decisions. 
In this paper, we propose a novel forecasting framework, named Self-adaptive Decomposed Interpretable framework~(SaDI), which ensembles long-term trend, short-term trend, and period modelings to capture temporal characteristics in different components. The external variable triggered loss is proposed for the imbalanced learning under extreme events. 
Furthermore, Generalized Additive Model (GAM) is employed in
the framework for desirable interpretability. The experiments on both Central China electric load and public energy meters from buildings show that the proposed SaDI framework achieves
average $22.14\%$ improvement compared with the current state-of-
the-art algorithms in forecasting under extreme events in
terms of daily mean of normalized RMSE. 
Code, Public datasets, and Appendix are available at: \url{https://doi.org/10.24433/CO.9696980.v1}.   
\end{abstract}

%% file: sections/1_introduction.tex
\section{Introduction} 
\vspace{-0.3cm}
\label{sec:intro}
The electric load forecasting (ELF) is one of the major problems facing the power industry~\cite{chan2012load,zhou2022robust}. Especially, when extreme events occur, load always fluctuates and threatens the electric grid. 
For example, China issued the highest heat alert for almost 70 cities in July 2022, and the electric load increased dramatically due to the extensive use of air conditioner. Thus, accurate forecasting under extreme events is highly desirable. 
Despite its importance, forecasting under extreme events is not well investigated. Modern deep learning based methods for time series forecasting~\cite{zhou2022fedformer,chen2022quatformer,nbeats/iclr2019} often focus on minimizing the global loss, which ignore data skew between normal cases and extreme events and fail to achieve desirable performance under extreme events.
Note that forecasting under extreme events is closely related to regression problems on imbalanced data, for which numerous methods have been proposed, such as SMOTER~\cite{IR/SMOTER2013}, SMOGN~\cite{IR/SMOGN2017}, reweighting~\cite{IR/reweighting}, transfer learning~\cite{IR/transfer_learning}, label distribution smooth (LDS)~\cite{IR_LDS/ICML21/}, etc. 
More related work can be found in Appendix \ref{app:related_work}.

To deal with load forecasting under extreme events, especially complicated load series mixed with long-term trend, short-term trend, and periodical patterns, we propose a novel framework named Self-adaptive Decomposed Interpretable framework (SaDI). It decomposes the original load series into three components, which are modeled differently.
We observe that the effects of extreme events caused by external covariables dominate load patterns. For example, the excessively high load in July 2022 in China is mainly caused by the high temperature. Thus, we further design an External Triggered Loss (ETL) to improve the forecasting performance.
In addition, interpretability is also an important factor for system operators~\cite{BuildingMachines/atal/2018/few_shot}.
We employ Generalized Additive Models (GAM)~\cite{lou2012intelligible,hastie1986generalized} to learn the explainable relationship between the short-term trend and input features, where GAM is a class of
intrinsic explainable methods that formulate the predicate function
as a summation of functions that only rely on single features~\cite{HowTrustworthyAreGams}. 
To summarize, our contributions are listed as follows:
\vspace{-0.2cm}
\begin{enumerate}
    \item The proposed SaDI is robust to extreme events thanks to its decomposed structure, and the decomposed series are treated with different strategies.
    \vspace{-0.2cm}
    \item The proposed SaDI is interpretable by adopting a Generalized Additive Model (GAM) for modeling the relationship between the target and input features.
    \vspace{-0.2cm}
    \item 
    We introduce a loss triggered by external variables (ETL), which further enhances the model with robust performance under extreme events. 
\end{enumerate}

%% file: sections/1.9_statement_of_problem.tex
\section{Statement of the Problem}

\begin{figure}[h]
    \centering
    \includegraphics[width=0.9\linewidth]{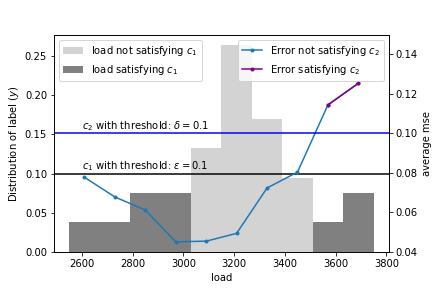}
    \vspace{-0.3cm}
    \caption{Illustration of how to define an $Extreme~Event$ with two conditions $c_1\ \&\ c_2$. $c_1$: the probability of the label is lower than a threshold $\epsilon$. $c_2$: the forecasting error is higher than a threshold $\delta$.}
    \label{fig:extreme_events_illustrate}
\vspace{-0.4cm}
\end{figure}

Extreme events are rare and random, but play a critical role in many real applications. 
In most cases, extreme events can be reflected by one or several indicators, either the label or the features.
In this paper, the dataset with a size of $N$ is represented as: $D=\{d_1,\dots,d_i,\dots,d_N\}$, where $d_i=(\boldsymbol{x_{i}},y_i)$, $\boldsymbol{x_{i}}$ is the input features of $i^{th}$ sample, $y_i$ is the  label. Then the subset of samples under extreme events $D_{ex}$ can be obtained by:
\begin{equation}
D_{ex}=\{ d_i|c(\boldsymbol{x_{i}},y_i),d_i\in D\},
\end{equation}
where $c(\cdot)$ is extreme events condition. 
For the construction of $c$, we assume that the extreme event happens when the label is rare, and at the same time, the prediction error is high. So we have $c = c_1 \cap c_2$, and: 

\[c_1:P(b_j)<\epsilon \] 
\[ c_2: \sum_{(\boldsymbol{x}_i,y_i)\in b_j} \|f(\boldsymbol{x}_i)-y_i \|_p > \delta\]
Here, we separate the dataset $D=\{(\boldsymbol{x}_i,y_i)\}$ into bins $\{b_j\}$. 
$c_1$ means that the probability of the labels $\{y_i\}$ in bin $b_j$ is less than $\epsilon$, which indicates the rareness of the extreme event. $c_2$ means that the prediction error (p-norm) in $b_j$ is more than $\delta$. $f(\cdot)$ is a baseline forecasting model, e.g., LightGBM or other deep learning methods. An illustration is depicted in Figure~\ref{fig:extreme_events_illustrate}.

%% file: sections/2_methodology.tex
\section{Methodology}

\subsection{Overall Framework}

\begin{figure}[t]
    \centering
    \includegraphics[width=0.7\linewidth]{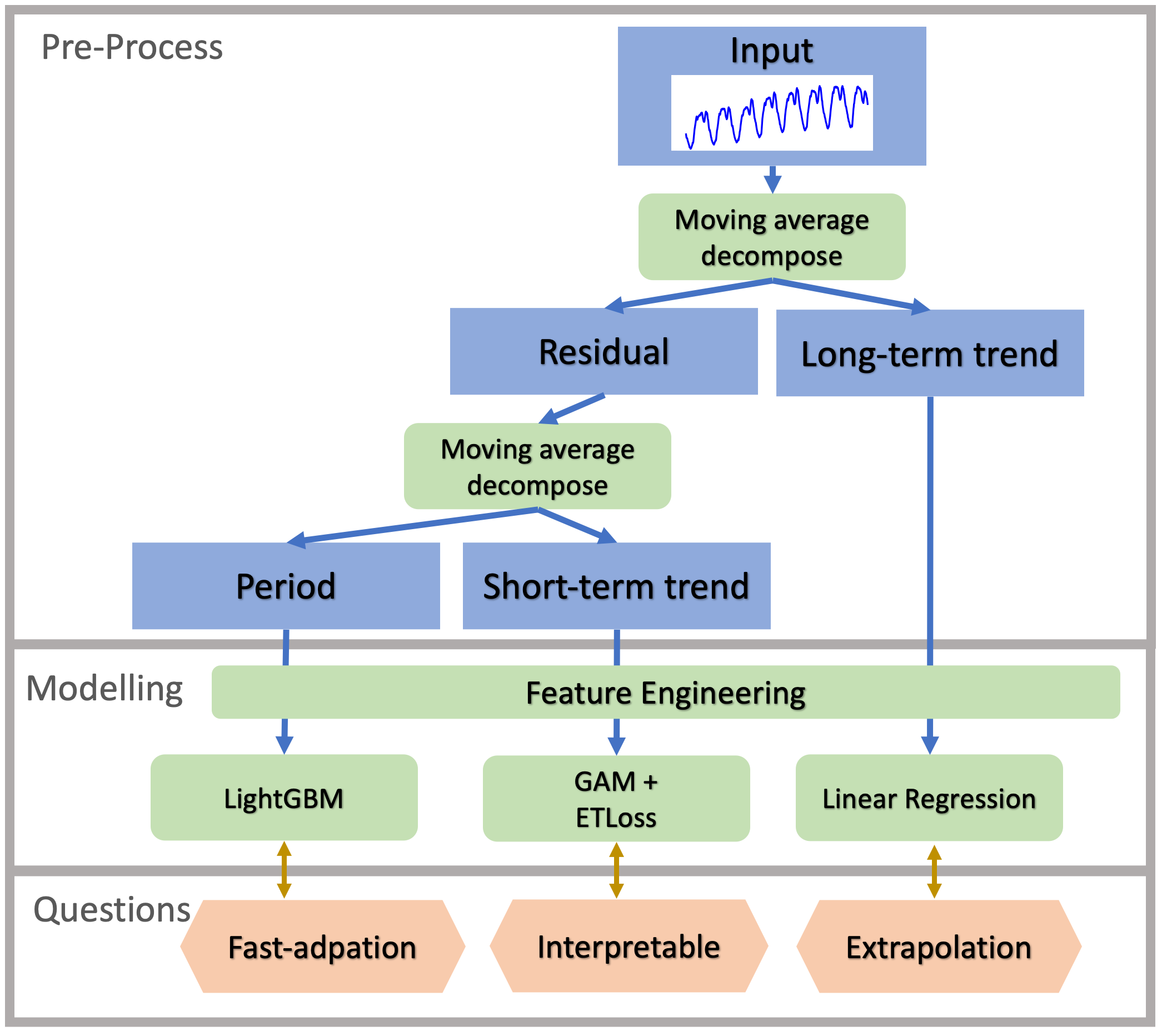}
    \vspace{-0.3cm}
    \caption{The overall framework of SaDI, which consists of decomposition-based pre-processing, feature engineering, and three models to deal with different decomposed components, respectively.}
    \label{fig:overall_framework}
\vspace{-0.4cm}
\end{figure}

The overall framework of our proposed model, SaDI, is an ensemble structure as shown in Figure~\ref{fig:overall_framework}. The input series is first processed by decomposition modules. Then we use Linear regression to model long-term trend, GAM with external variable triggered loss to model short-term trend, and LightGBM to model period.


\subsection{Decomposition-based Modeling}
\begin{figure*}[t]
    \centering
    \includegraphics[width=0.7\linewidth]{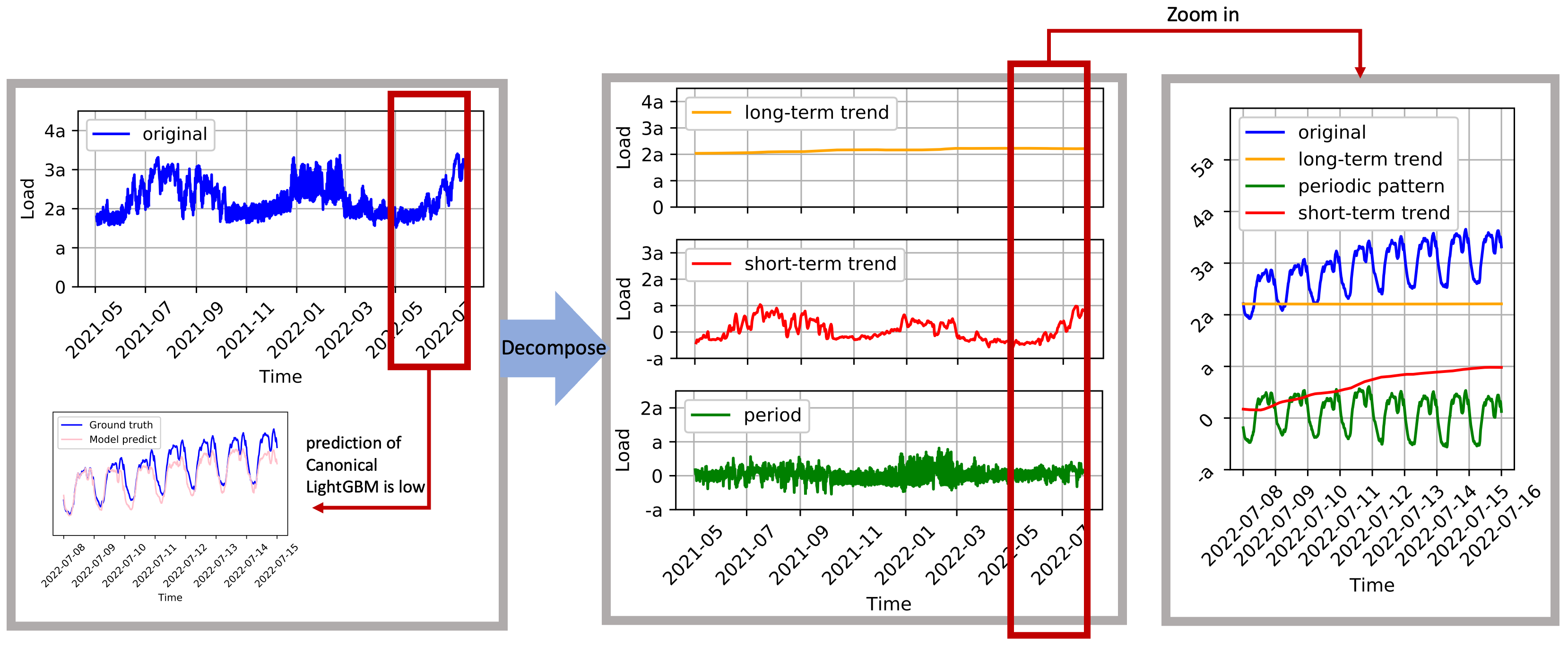}\vspace{-3mm}
    \caption{Decomposition of electric load series. (Left up): The original series. (Left down) The prediction given by canonical LightGBM is low under extreme events. (Mid): Decomposition of original load into three components: long-term trend, short-term trend, and period. (Right): zoom in on the three components. The Y axis is masked for confidentiality purposes.}
    \label{fig:decompose_framework}
\vspace{-0.3cm}
\end{figure*}

The electric load series are often a mixture of trends and periods. We decompose the electric power load ($y_t$) into a long-term trend ($y_t^{LT}$), a short-term trend ($y_t^{ST}$), and a periodic component ($y_t^S$) as: $y_t = y_t^{LT} + y_t^{ST} + y_t^{S}$. We adopt a moving-average-based decomposition method defined as:
\begin{equation}
    y_t^{LT} = MovAvg(y_t), R = y_t - y_t^{LT},
\end{equation}
\begin{equation}
    y_t^{ST} = MovAvg(R), y_t^S = R - y_t^{ST}.
\end{equation}
The decomposed components are illustrated in Figure~\ref{fig:decompose_framework} (Mid). Three sub-series estimations are discussed below.

\textbf{Long-term Trend Modeling}
The long-term trend, also known as the yearly trend, is a smooth and continuously growing curve as shown in Figure~\ref{fig:decompose_framework} (Mid). The tendency of electric load series strongly correlates to the growth of district GDP or climate change, such as global warming~\cite{Survey_classification/alfares2002electric}.
We use linear regression which is capable of handling the simple pattern of long-term trend while maintaining the ability of extrapolation on rising tendency. 



\textbf{Short-term Trend Modeling}
After removing the long-term trend, 
the residual combines periodical-term and short-term trend.
In short-term trend, due to the rareness of extremely hot weather in training data, a sudden increase in load would be difficult for the model to capture. 
To solve this, we introduce an external variable triggered loss (ETL) which will be discussed in Section \ref{sec:evl}. We leverage GAM as a backbone model in Section \ref{sec:GAM} for interpretability with ETL.  

\textbf{Period Modeling}
After removing the long-term trend and the short-term trend, We use LightGBM to model the periodic daily pattern without long-term and short-term trend. 

\subsection{External-variable Triggered Loss}
\label{sec:evl}

In our framework, after decomposition, the short-term trend is affected by external variables like weather indicators. To quantitatively characterize the effects of the external variables, an External-variable Triggered Loss (ETL) function is designed as:
\begin{align}
    ETL =   \sum_{t=1}^N\left[{S\left(\sum_{q=1}^Q \lambda_{q} {x_{t,q}^e}\right)*(\hat{y_{t}}-y_t^{ST})^2}\right],\label{loss:etl}
\end{align}
where $x^e$ denotes the external variable, $N$ denotes the number of samples, $\lambda_q$ denotes the user-defined weight for the $q^{th}$ external features $x_q^e$ (for example, in load forecasting, we give large weights for temperature), and $\hat{y}_t$ denotes the output of GAM. $S(\cdot)$ is a non-linear score function, which gives different weight to each sample according to its extreme level.

We now discuss the selection of weights $\{\lambda_q\}$ and score function $S(\cdot)$. 
we set a different weight to selected external variables by correlation analysis when extreme value occurs. Specifically, when scorching heats happened, temperature is selected as an external variable as a result of high correlation coefficient with target load. It can also be explained physically that residential usage of air conditioning load would increase significantly when air temperature increases. 
In this simple case, temperature is the only external variable. Let $X_{t,1}^e$ denote the feature that represents temperature, we set $\lambda_1=1$
and $\lambda_j=0,\ \forall j\ne 1$.  To emphasize the weight of samples with extremely high temperatures, we set $S(\cdot)$ as
\begin{equation}
S(T_t)=
\begin{cases}
\frac{2}{ 1+e^{{-(T_t-K)}})}& T_t>K, \\
1& T_t\le K,
\end{cases}
\end{equation}
where $T_t=\lambda_1X_{t,1}^e$, $S(.)$ is a piece-wise function adopted to give heavier weights to samples with temperatures higher than $K$, and $K$ is set as a predefined threshold value for different tasks. Multi external variables ETL expression can be easily expanded with \ref{loss:etl}.

\subsection{Generalized Additive Model (GAM)}
\label{sec:GAM}
\begin{figure}[t]
    \centering
    \includegraphics[width=1.0\linewidth]{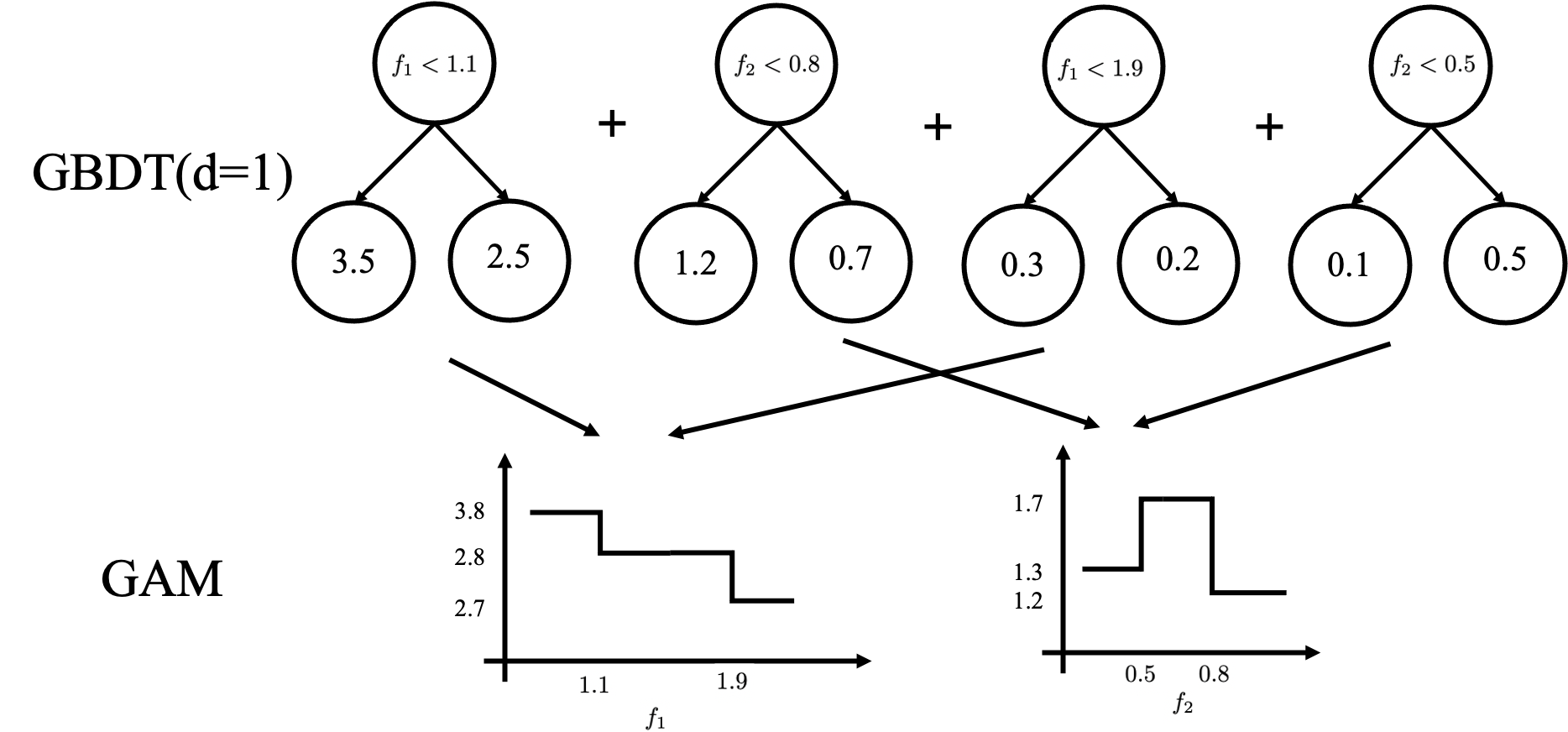}\vspace{-3mm}
    \caption{Example of converting GBDT to GAM.}
    \label{fig:gbdt_gam}\vspace{-4mm}
\end{figure}

As discussed above, the short-term trend is correlated to external factors. We leverage a GAM model to fit the short-term trend with ETL as loss function, where the short-term trend is formalized as a summation of univariate functions of the external factors. Specifically,
\begin{align}
y_t^{ST} = \phi_0+\sum_{q=1}^{Q}\phi_q(x^e_{t,q}) + \xi_{t},\label{gam}
\end{align}
where $\phi_q(\cdot)$ denotes the function of the $q^{th}$ external variable,
and $\xi_t$ denotes the fitting error. 
GAM has great interpretability. Given a sample $(\boldsymbol{x}_t,y_t)$, we can readily get the contribution of each factor
$x^e_t$ as $\phi_q(x^e_t)$.
Obviously, the set of $\{\phi_q\}$ that satisfies (\ref{gam}) is not unique.
Some works select $\{\phi_q\}$ from the space
formed by some particular basis (such as B-spline basis)~\cite{hastie2017generalized,pierrot2011short}.
Here we implement GAM using GBDT (Gradient Boosting Decision Tree) 
by setting the depth of trees in GBDT to 1 as shown in Figure~\ref{fig:gbdt_gam}, which means each tree in only uses one feature, thus no feature interaction is involved.

%% file: sections/3_experiments.tex
\input{tables/tab_benchmark}

\begin{figure*}[t]
\vspace{-0.2cm}
    \centering
    \begin{subfigure}[t]{0.3\textwidth}
        \centering
        \includegraphics[width=\textwidth]{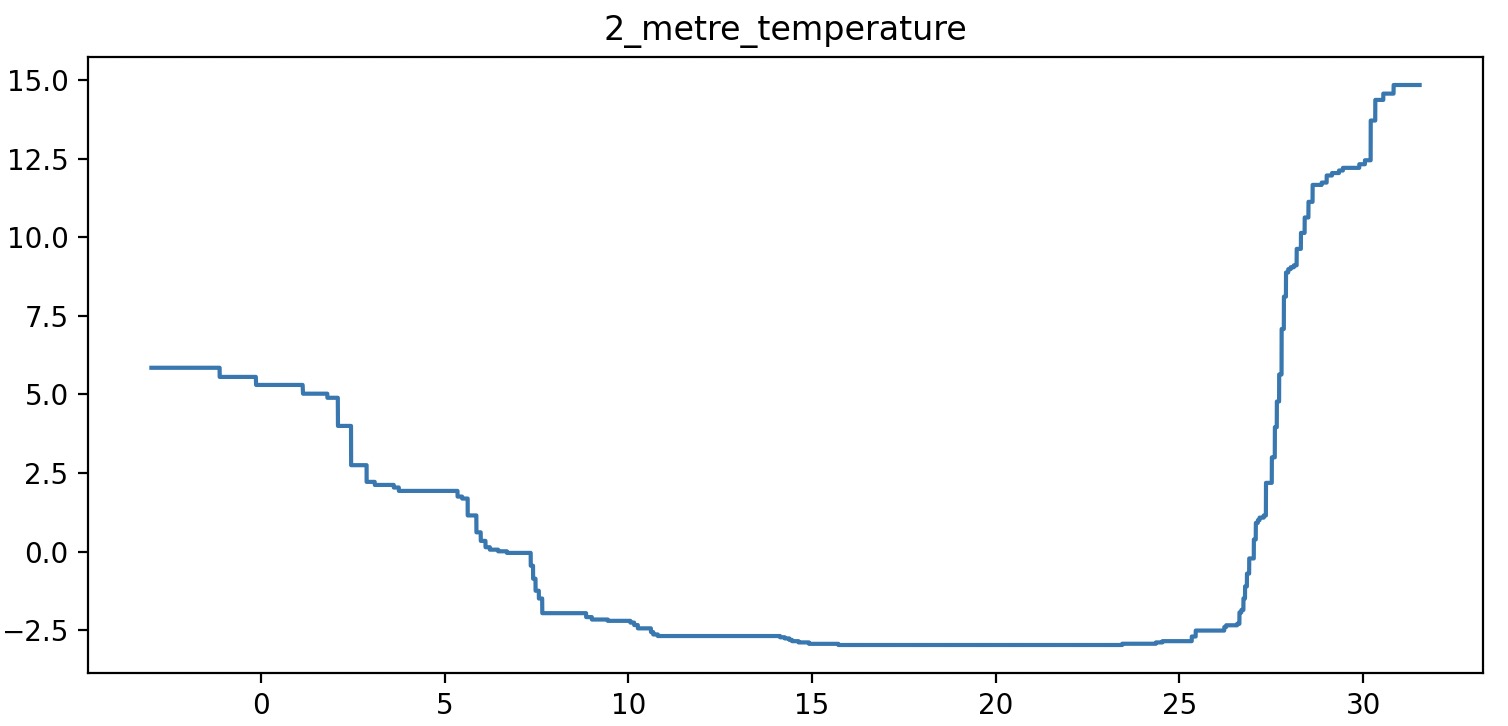}
        \caption{Temperature.}
        \label{fig:temperature}
    \end{subfigure}
    \hfill
    \begin{subfigure}[t]{0.3\textwidth}
        \centering
        \includegraphics[width=\textwidth]{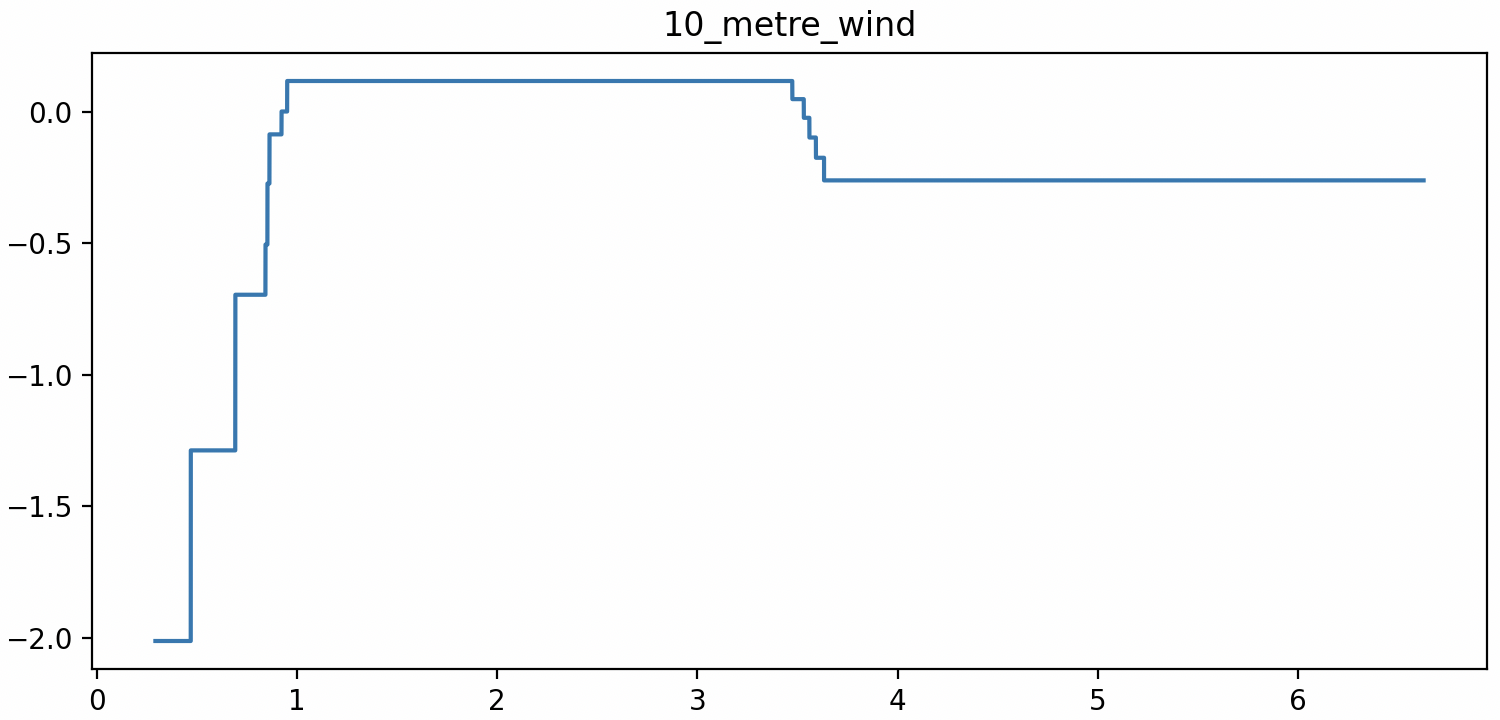}
        \caption{Wind speed.}
        \label{fig:wind}
    \end{subfigure}
    \hfill
    \begin{subfigure}[t]{0.33\textwidth}
        \centering
        \includegraphics[width=\textwidth]{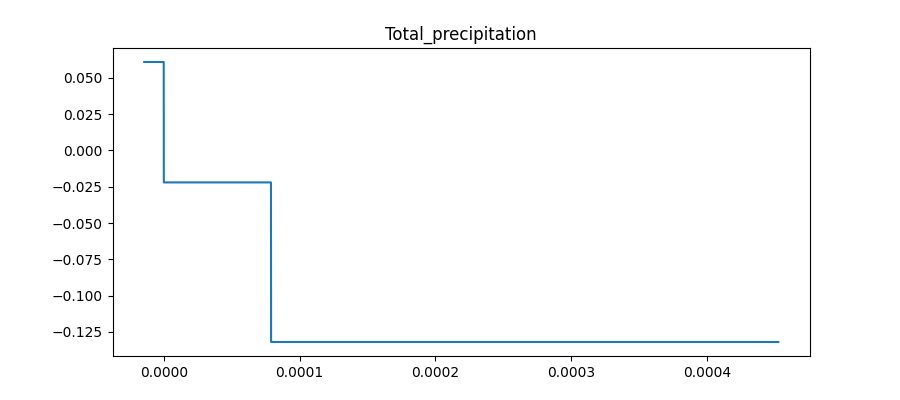}
        \caption{Total precipitation.}
        \label{fig:total_precipitation}
    \end{subfigure}
    \vspace{-3mm}
    \caption{Learnt functions for temperature, wind speed, and whether special holiday by SaDI.}
    \label{fig:gam}
\vspace{-4mm}
\end{figure*}

\section{Experiments}




\subsection{Datasets and Baselines}
\vspace{-0.2cm}
Two electric load datasets are introduced to evaluate our proposed framework. The private datasets are real-world load data from Mid-centre China. 
The public datasets are from the {ASHRAE} Great Energy Predictor {III} competition~\cite{miller2020ashrae}, like Peacock, Rat, and Robin. 
We compare the performance of SaDI and baselines including LightGBM~\cite{ke2017lightgbm}, LSTM~\cite{hochreiter_long_1997_lstm}, N-BEATS~\cite{nbeats/iclr2019}, TCN~\cite{tcn_electric/2021}, EVT~\cite{EVLoss/kdd19/}, LDS~\cite{IR_LDS/ICML21/}. 
More details about the datasets, baselines, and feature engineering can be found in Appendix \ref{app:dataset}, Appendix \ref{app:baselines}, and Appendix \ref{app:feature_engineering}, respectively.

\subsection{Evaluation Metrics}
\vspace{-0.3cm}
The most widely used metrics in forecasting are RMSE and MAPE. The
RMSE is scale-dependent and unsuitable for comparing
forecasting results at different aggregation levels. We adopt the daily mean of normalized root mean squared error $\rm{nRMSE_d}$ and $\rm{MAPE_d}$.  
Their definitions are
$
  \rm{nRMSE_d} = \frac{1}{N}\sum_{n=1}^N \left(\sqrt{\frac{1}{M}\sum_{i=1}^M
    (\frac{y_i-\hat{y}_i}{y_i})^2} \right)
$
and $\rm{MAPE_d} = \frac{1}{N}\sum_{n=1}^N \left(\frac{1}{M}\sum_{i=1}^M \frac{|y_i-\hat{y}_i|}{y_i} \right) $, 
where $M$ is the number of points in one day (normally $M$=96, i.e. the sampling interval is 15 minutes), and $N$ is the number of days to be evaluated.

\subsection{Performance Comparisons}
\vspace{-0.3cm}

\begin{figure}[t]
\vspace{-0.4cm}
    \centering
    \includegraphics[width=0.8\linewidth]{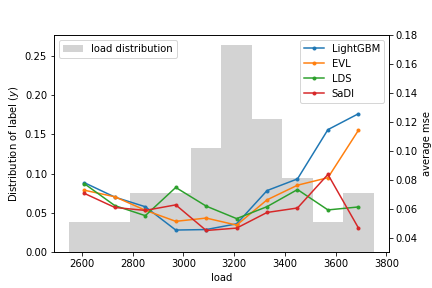}
    \vspace{-0.4cm}
    \caption{Performance comparison of SaDI with baselines under the different bins of loads.
    }
    \label{fig:distribution_compare_mse}
    \vspace{-5mm}
\end{figure}

The results of the baselines and the proposed SaDI  are summarized in Table \ref{tab:benchmarks}. The baselines can be categorized as tree-based models (LightGBM) and deep learning models (N-BEATS, TCN, and LSTM). SaDI, LightGBM, EVL, and LDS share the same feature engineering process. 
It is observed that SaDI achieves the best results in terms of two metrics on almost all datasets. 
Specifically, SaDI improves the $\rm{nRMSE_d}$ and $\rm{MAPE_d}$ metrics on average by $20.0\%$ and $22.14\%$, respectively, compared with the best baseline (LDS). 
Furthermore, Figure \ref{fig:distribution_compare_mse} shows performance of SaDI on bins of loads compared with the baselines mentioned above. In extreme events when loads exceed 3600 or are below 2800, the RMSE of SaDI is lower than other baselines.

\subsection{Interpretability}
\vspace{-0.2cm}
As GAM is employed as a part of SaDI, we can exploit the inherent property of GAM to explain the forecasting of SaDI. As an illustration, the effect on load as functions of
temperature, wind speed, and total precipitation by SaDI are plotted in Figure~\ref{fig:gam}. For example,
Figure~\ref{fig:temperature} shows how each value of temperature contributes differently to the overall energy load. 
When the temperature exceeds 26 or is below 7.5 Celsius degree, the energy load changes dramatically. 



\subsection{Other Experiments}
\vspace{-0.3cm}
For more experiments (visualization of curves, ablation study, evaluation of speed), please refer to Appendix \ref{app:experimental_res}.

%% file: tables/tab_benchmark.tex
\begin{table*}[t]
\centering
\begin{footnotesize}
\caption{Performance comparison of load forecasting on two real-world datasets. The best performance is highlighted in bold. Lower values of $nRMSE_d$ and $MAPE$ indicate better performance.}\vspace{-3mm}
\scalebox{0.7}{
\begin{tabular}{c|c|cc|cc|cc|cc|cc|cc|cc|cc|cc}
\hline
\multicolumn{2}{c|}{Methods}&\multicolumn{2}{c|}{\textbf{Proposed SaDI}}&\multicolumn{2}{c|}{LightGBM}&\multicolumn{2}{c|}{N-BEATS}&\multicolumn{2}{c|}{TCN}&\multicolumn{2}{c|}{LSTM}&\multicolumn{2}{c|}{LDS*}&\multicolumn{2}{c|}{EVL}\\
\hline
\hline
\multicolumn{2}{c|}{Datasets} & $nRMSE_d$ & $MAPE_d$ & $nRMSE_d$  & $MAPE_d$ & $nRMSE_d$  & $MAPE_d$ & $nRMSE_d$  & $MAPE_d$ & $nRMSE_d$  & $MAPE_d$ & $nRMSE_d$  & $MAPE_d$ & $nRMSE_d$  & $MAPE_d$\\
\hline






\multirow{4}{*}{Huazhong} & Hubei & \textbf{0.052} & \textbf{0.045} & 0.066 & 0.058 & 0.106 & 0.099 & 0.075 & 0.067 & 0.112  & 0.114 & 0.061 & 0.054 & 0.064 & 0.054 \\

&Hunan & 0.046 & \textbf{0.039} &  \textbf{0.044} & 0.047 &  0.123 & 0.118 & 0.068 & 0.061 & 0.133 & 0.121 & 0.054 & 0.047& 0.055&0.047 \\

&Henan & \textbf{0.059} & \textbf{0.051} &  0.082 & 0.072 & 0.105 & 0.097 & 0.099 & 0.090 & 0.128 & 0.114 & 0.079&0.069 & 0.080 & 0.069     \\

&Jiangxi& \textbf{0.044} & \textbf{0.038} & 0.053 & 0.045 & 0.071 & 0.064 & 0.078 & 0.060 & 0.088 & 0.080 & 0.057 & 0.051 & 0.057 & 0.048\\
\cline{1-2}
\cline{1-2}
\multirow{3}{*}{Public} 
& Peacock & \textbf{0.025} & \textbf{0.021} & 0.058 & 0.046 & 0.060 & 0.049 & 0.041 & 0.034 & 0.053 & 0.048 & 0.032 & 0.028 & 0.040 & 0.033\\
& Rat & \textbf{0.086} & \textbf{0.077} & 0.163 & 0.152 & 0.174 & 0.157 & 0.130 & 0.121 & 0.194 & 0.183 & 0.130 & 0.121 & 0.156 & 0.147\\
& Robin & \textbf{0.068} & \textbf{0.060} & 0.115 & 0.101 & 0.080 & 0.071 & 0.075 & 0.067 & 0.198 & 0.190 & 0.077 & 0.071 & 0.073 & 0.065\\
\hline
\end{tabular}
\label{tab:benchmarks}
}
\end{footnotesize}
\end{table*}

%% file: sections/4_conclusions.tex



\vspace{-3mm}
\section{Conclusions}
\vspace{-3mm}
In this paper, we propose a Self-adaptive Decomposed Interpretable (SaDI) framework for electric load forecasting under extreme events. Our framework decomposes load into long-term, short-term, and period patterns, and deals with them separately with corresponding models. To improve sensitivity to external variables under extreme events, an external variable triggered Loss is designed to guide forecasting models. Furthermore, To explain the forecasting results, generalized additive models are incorporated to provide each feature's contribution to the predicted values quantitatively.

%% file: appendix/1_appendix.tex
\newpage

\appendix

\section{Supplementary material Introduction}
\begin{figure}[h]
    \centering
    \includegraphics[width=0.9\linewidth]{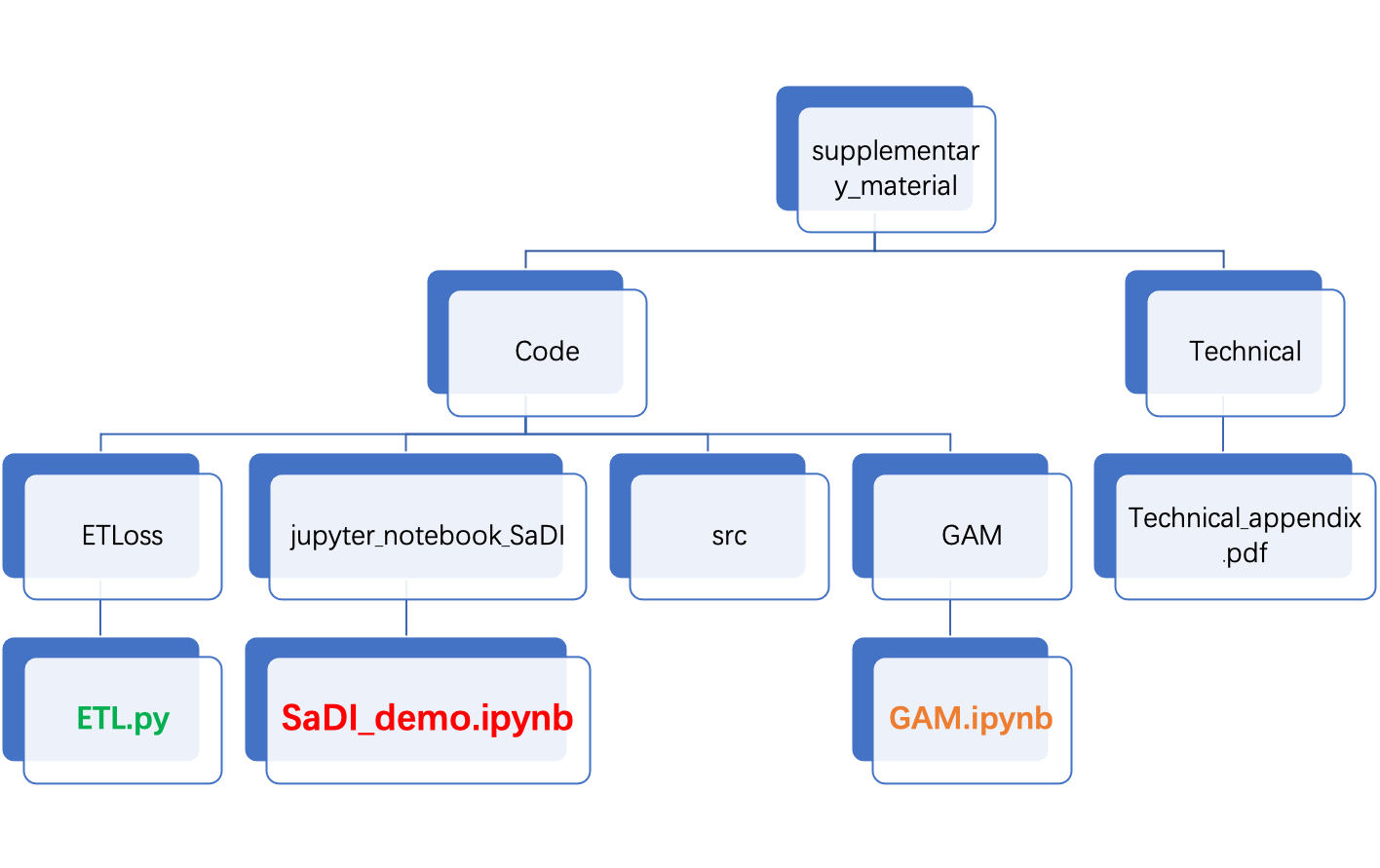}
    \vspace{-.2cm}
    \caption{material introduction}
    \label{figure 1: material introduction}
\end{figure}

The appendices are all in "supplementary\_materials" folder.
In "Code" Folder, "SaDI\_demo.ipynb"(red in Figure~\ref{figure 1: material introduction}) is the main entrance function of SaDI framework. sub-folder "ETL.py" (green in Figure~\ref{figure 1: material introduction}) gives the definition referred at section "External-variable Triggered Loss" in main paper.
"GAM.ipynb" (orange in Figure~\ref{figure 1: material introduction}) in "GAM" folder provides source code to generate visible explainable figures. 
In "Technical" Folder, this "Technical\_appendix.pdf" describes appendix materials and illustrates the "Experiments" procedure in the main paper.

\input{sections/1.5_related_work.tex}

\section{Experiment Details}

\subsection{Datasets}
\label{app:dataset}
Two Data sets are provided to verify our framework works well. The First data set is private from practical load data from Mid-centre China and South-East China. The Second data is public from the {ASHRAE} Great Energy Predictor {III} competition.

First, we use a large-scale private ELF datasets: Huazhong (HZ). HZ dataset contains four sub-datasets for four districts (Hubei, Hunan, Henan, and Jiangxi) in Central China. For each district, the sub-dataset contains one series of electric load and 14 covariates, indicating weather conditions in the future. All time series are sampled with an interval of 15-minutes. Other details of the data set can be found in Table~\ref{tab:dataset}.

Secondly, public data set "The Building Data Genome 2"(BDG2) is an open data set made up of 3,053 energy meters from 1,636 buildings. The time range of the times-series data is the two full years and the frequency is hourly measurements of electricity, heating and cooling water, steam, and irrigation meters. These meters were collected from 19 sites across North America and Europe, with one or more meters per building measuring whole building electrical, heating and cooling water. After grouping by 19 sites, related weather data and cleaned load data are provided for energy forecasting. Other details of the data set can be found in Table~\ref{tab:dataset}.
\input{tables/tab_dataset}

\subsubsection{Data Confidential Statements}
Raw load data produced by Central-China and South-east China 's grid company is confidential constraint by agreements. Regretly, we could not provided related data in public. The Y axis is masked for confidentiality purposes in both main paper and appendix. However, the Experiments procedure will illustrate here in details to help reader to understand our works.


\subsection{Baselines}
\label{app:baselines}

We compare the performance of SaDI with 6 baselines, 
including four general time series models and two models specifically designed for extreme events.
\begin{enumerate}
    \item LightGBM~\cite{ke2017lightgbm}: A tree-based boosting model, feature engineering required. 
    \item LSTM~\cite{hochreiter_long_1997_lstm}: Variants of recurrent neural networks, capable of capturing long-term dependency.
    \item TCN~\cite{tcn_electric/2021}: Temporal convolutional network, convolution-based time series model.
    \item N-BEATS~\cite{nbeats/iclr2019}: Fully connected structure with backward and forward residual links, a strong SOTA.
    \item EVL~\cite{EVLoss/kdd19/}:  Extreme value loss, which is proposed from EVT, provides better predictions of extreme events.
    \item LDS~\cite{IR_LDS/ICML21/}: Deep imbalanced regression model learned from extremely imbalanced data with continuous targets, calibrating target distribution with label distribution smoothing (LDS). 
\end{enumerate}

Note that we simplify EVL referred in \cite{EVLoss/kdd19/} as only extremely high values events. With the following formulation, $u$ represents the extreme degrees of load,
\begin{equation}
u=
\begin{cases}
\frac{\hat{y}-\epsilon}{\hat{y}}& \hat{y}>\epsilon \\
0& \hat{y}\le\epsilon
\end{cases}
\end{equation}
where $\epsilon=\mu+K\sigma$ gives the upper bound, with $\mu, \sigma$ being the mean and standard deviation of load.
Then, $EVL(\cdot)$ designed in~\cite{EVLoss/kdd19/} is used in $L_1=\sum_{t=1}^T{||\hat{y_t}-y_t||}+\lambda_1 EVL(u)$.
The gradient and hessian of $L_1$ can be easily obtained, and subsequently guide tree-base models, such as XGB, LightGBM, with customized loss.

\section{Feature Engineering}
\label{app:feature_engineering}

\begin{table*}[h]
\centering
\caption{Features after feature engineering}
\label{tab:feature_engineering}
\begin{footnotesize}


\resizebox{0.8\linewidth}{!}{ 
\begin{tabular}{c|c|c|c}
\hline
{Temporal features}&\multicolumn{1}{c|}{NWP features}&\multicolumn{1}{c|}{Load\_Rolling features}&\multicolumn{1}{c|}{Difference features}\\
\hline
\hline
\textbf{year} &\textbf{2\_metre\_temperature} & \textbf{load\_win\_7\_offset\_192\_median}&
\textbf{2\_metre\_temperature\_diff\_offset\_192} \\

\textbf{month} & \textbf{Surface\_pressure} & \textbf{load\_win\_7\_offset\_192\_mean}&
\textbf{Surface\_pressure\_diff\_offset\_192}\\

\textbf{day} & \textbf{Total\_cloud\_cover} & \textbf{load\_win\_7\_offset\_192\_min}&
\textbf{Total\_cloud\_cover\_diff\_offset\_192}\\

\textbf{is\_workday} & \textbf{Total\_precipitation} & \textbf{load\_win\_7\_offset\_192\_max}&
\textbf{Total\_precipitation\_diff\_offset\_192}\\

\textbf{is\_holiday} & \textbf{Skin\_temperature} & \textbf{load\_win\_7\_offset\_192\_std}&
\textbf{Skin\_temperature\_diff\_offset\_192}\\

\textbf{is\_weekend} & \textbf{...} & \textbf{load\_win\_7\_offset\_192\_skew}&
\textbf{...}\\

\textbf{day\_of\_month\_sin} & \textbf{...} & \textbf{load\_win\_7\_offset\_192\_q025}&
\textbf{...}\\

\textbf{...} & \textbf{...} & \textbf{...}&
\textbf{...}\\

\hline
\end{tabular}
}
\end{footnotesize}
\end{table*}

\begin{figure}[h]
    \centering
    \includegraphics[width=0.8\linewidth]{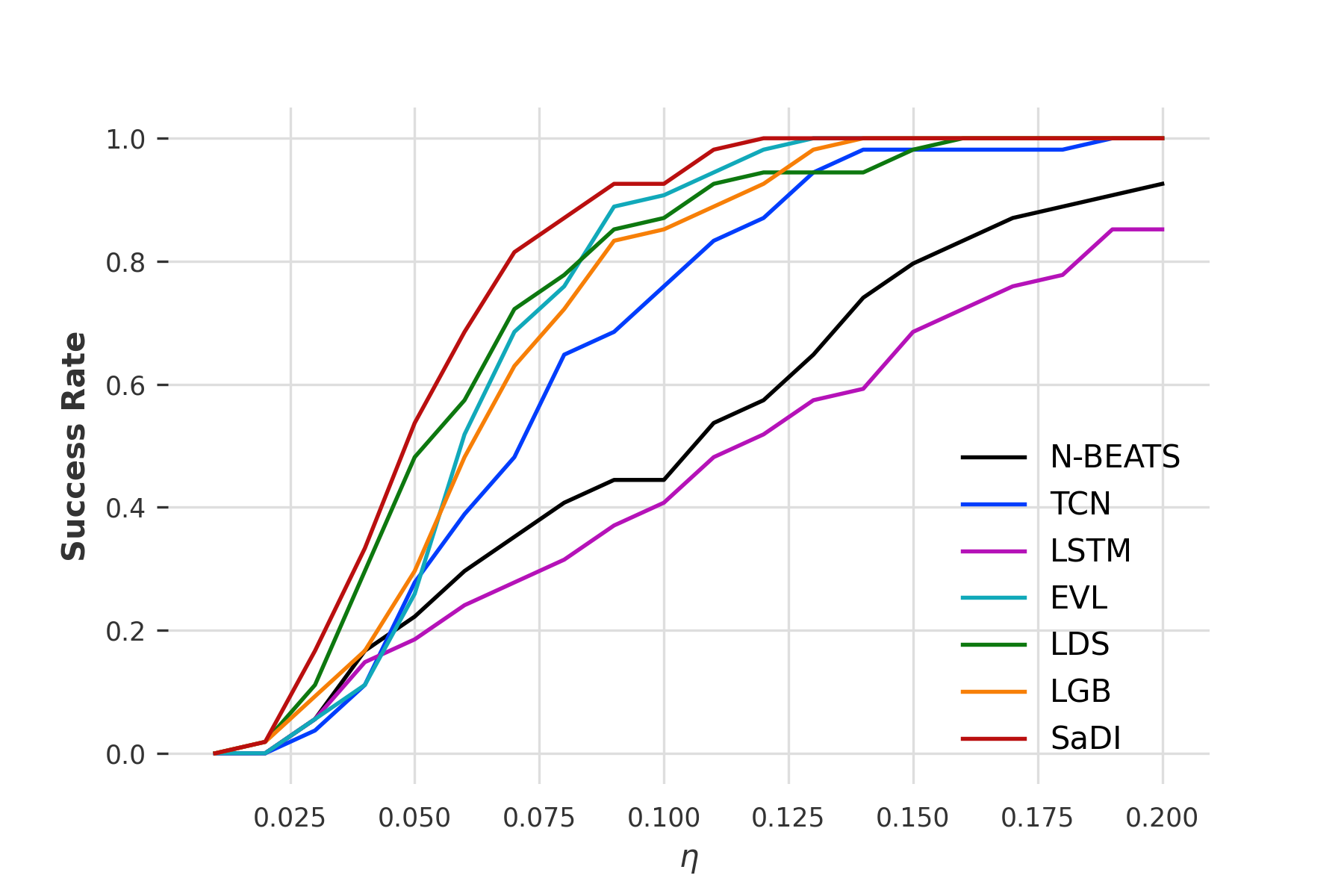}
    \vspace{-.2cm}
    \caption{Success rate with different threshold $\eta$ of SaDI and baselines on dataset of Hubei (55 days).}
    \label{fig:success_rate}
\end{figure}

Original numerical weather prediction (NWP) factors and history load are main attributes before engineering.
It includes 14 NWP attributes, such as "2 metre temperature"," surface pressure", "total cloud cover", "total precipitation","skin temperature",etc. History load in time series every 15min is also provided for engineering.

\begin{figure}[h]
    \centering
    \includegraphics[width=0.9\linewidth]{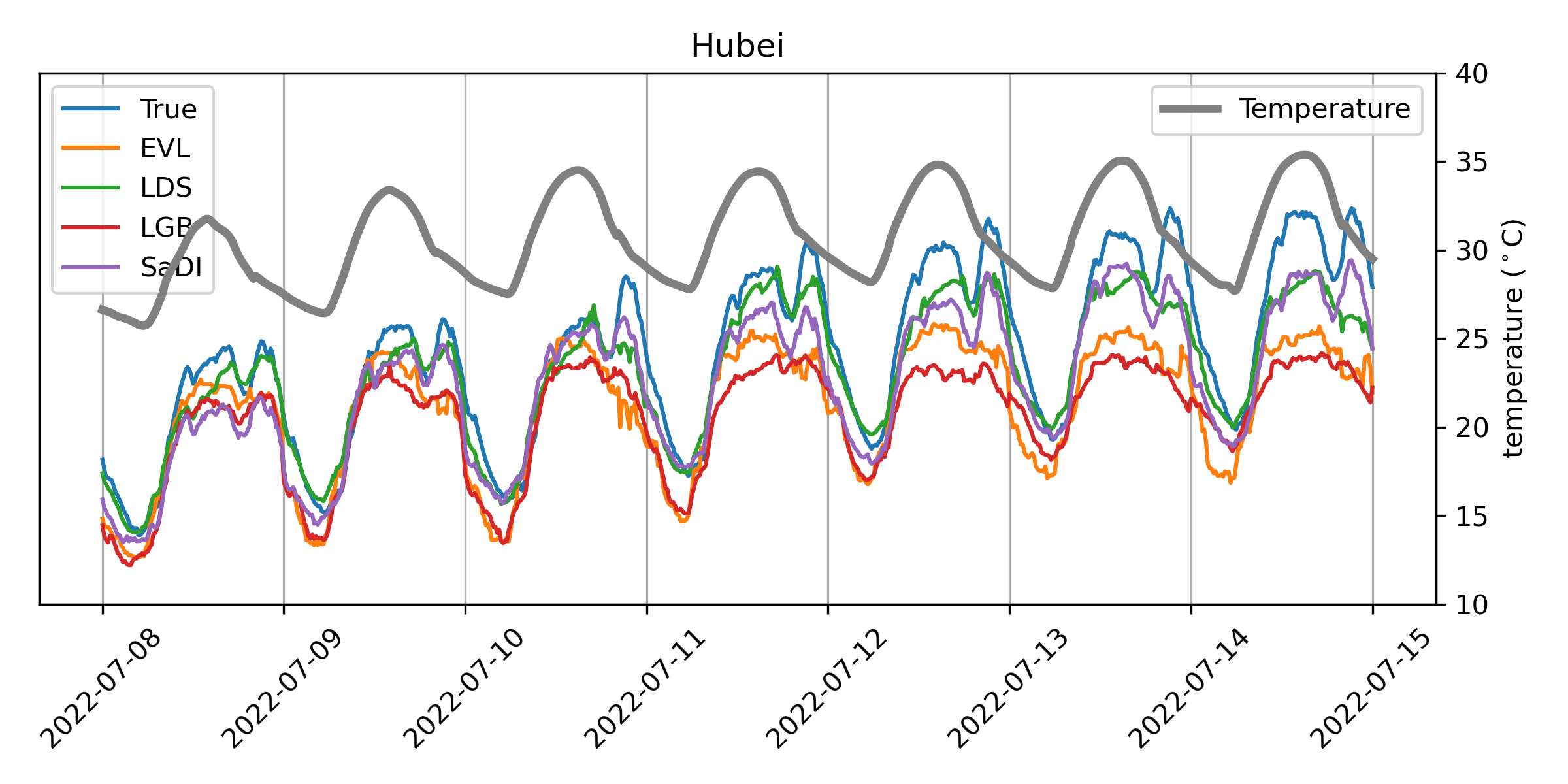}
    \includegraphics[width=0.9\linewidth]{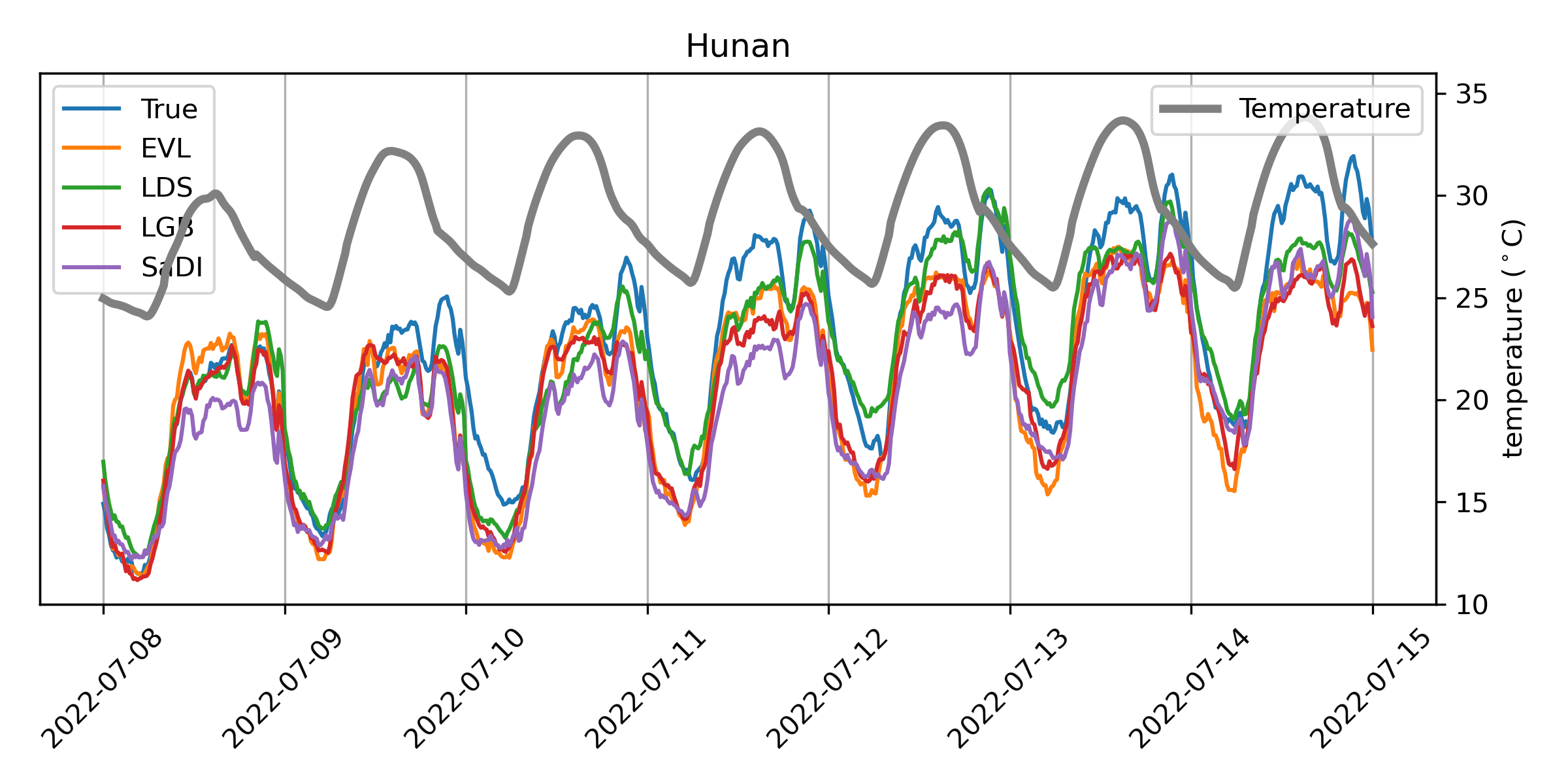}
    \includegraphics[width=0.9\linewidth]{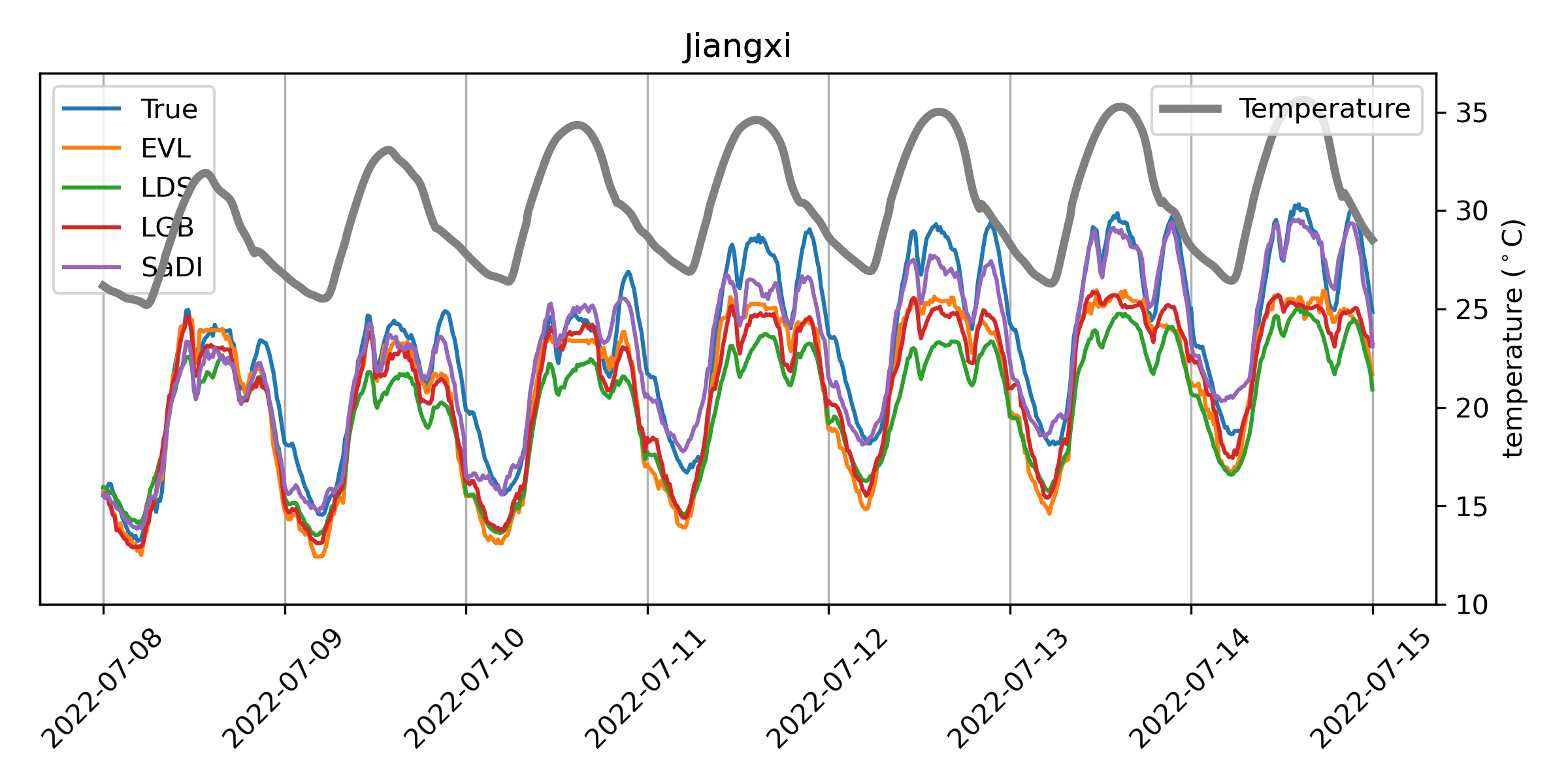}
    \includegraphics[width=0.9\linewidth]{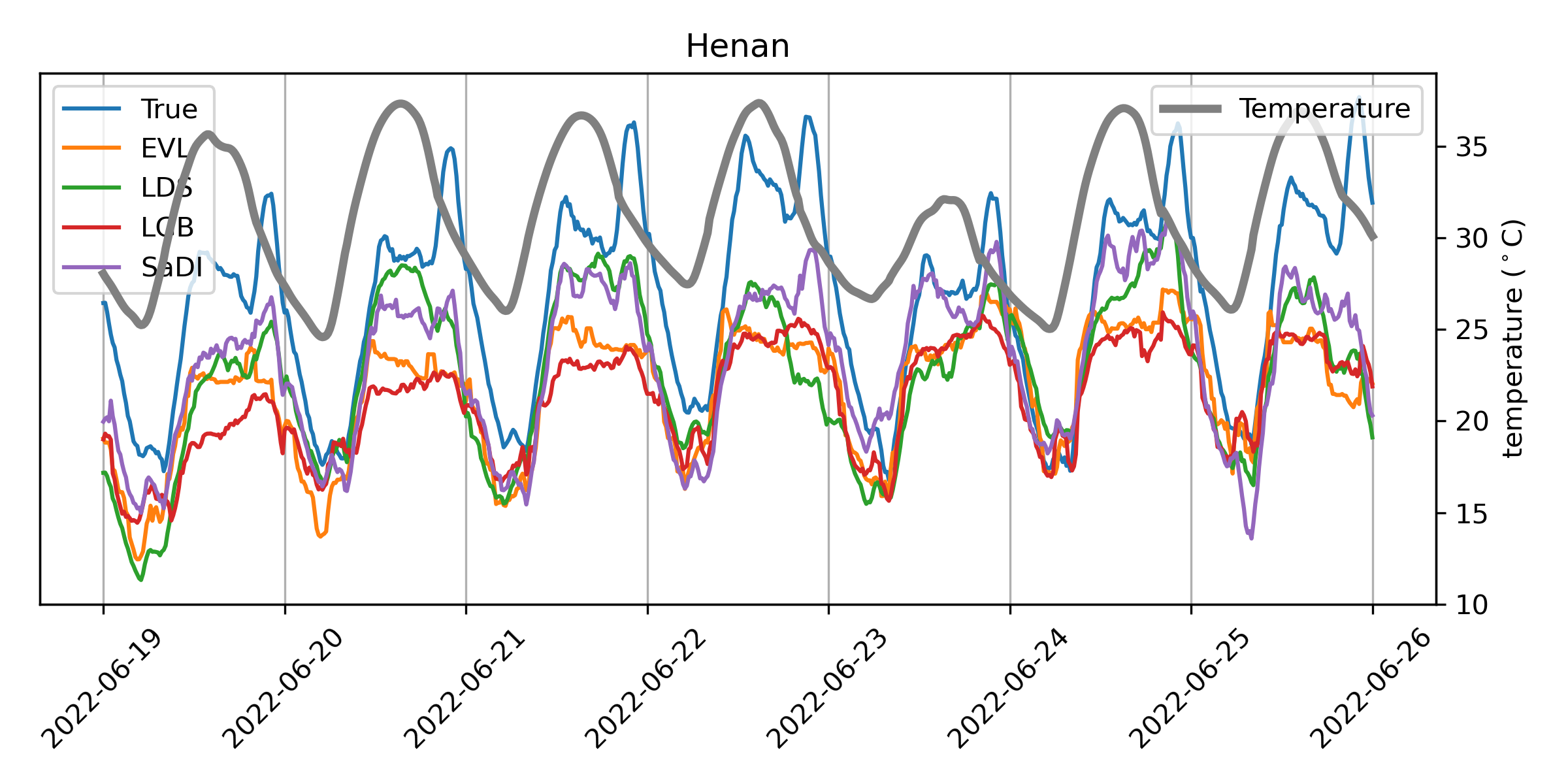}
    \caption{Performance comparison of SaDI with canonical LightGBM model and two baselines (EVT and LDS) designed for dealing with extreme events. We select two 7-day extreme events on the datasets of Hubei (up) and Hunan (down). The Y axis is masked for confidentiality purposes.}
    \label{fig:curve_benchmark}
\end{figure}

After feature Engineering described in "SaDI\_demo.ipynb", show in Table \ref{tab:feature_engineering} below, four categories features: Temporal features, NWP features, Load rolling features, Difference features, Rolling features, totally 63 are generated.
For load rolling features, $load\_win\_\{i\}\_offset\_\{j\}\_\{agg\}$ means rolling $15j$ minutes before with window size $i$ under aggregation method $agg$, such as $mean$,$max$,$min$,$skew$,etc.
For Difference features, $attribute\_diff\_offset\_\{j\}$ means making NWP $attribute$ difference operator with  $15j$ minutes before.

\section{Pseudocode} 
We summarize our self-adaptive, decomposed, and interpretable framework for electricity load
forecasting under extreme events
(referred to as SaDI)
in Algorithm 1.

\input{tables/algorithm1}

\input{tables/main_procedures}

\section{Experimental Results}
\label{app:experimental_res}

\subsection{Visualization of the Predicted Curve}
\label{app:extrme_events_curve}

In Figure \ref{fig:curve_benchmark} we show the curve predicted by SaDI and baselines. Four sub-cases (Hubei, Hunan, Jiangxi, and Henan) in the dataset of Huazhong are illustrated. We select a window of 7-day extreme event. The Y axis is masked for confidentiality purposes.

As shown in Figure~\ref{fig:curve_benchmark}, we visualize the curve predicted by models during extreme events. It is easy to observe that canonical LightGBM model failed to capture the temporal growth of load due to difficulty in extrapolation for tree-based model as we mentioned above, especially for the unseen peak in history. Introducing EVL into LightGBM makes the prediction better, but the gap between the prediction and the ground truth is still remarkable. LDS has comparable performance with SaDI, however, being a TCN-based model, its computational cost is much higher.

\subsection{Comparison of `Success Rate'}


Define $\boldsymbol{y}=[y_0,\dots,y_{m-1}]$,
$\boldsymbol{x}_t^h=[x_{t,0}^h,\dots,x_{t,p-1}^h]$,
$\boldsymbol{x}_t^e=[x_{t,0}^e,\dots,x_{t,q-1}^e]$,
and $\boldsymbol{x}_t=[(\boldsymbol{x}^h_t)^T,(\boldsymbol{x}^e_t)^T]^T$.
Here, we aim to find a function $f(\cdot)$ such that the success rate of prediction
is maximized. A prediction is said to be successful if the normalized root mean square error (nRMSE) is less than a predefined threshold $\eta$.
We denote the success rate with threshold $\eta$ as 
$\text{SR}@\eta$.
Then, given a set of
$\{(\boldsymbol{x}_t,y_t)\}$, we aim
to maximize:
\begin{align}\label{equ:success_rate}
    \text{SR}@\eta=\mathbb{E}\left[
    \mathcal{I}\left(\sqrt{\frac{1}{m}\sum_{t=0}^{m-1}\frac{(y_t-f(\boldsymbol{x}_t))^2}{y_t^2}}\leq\eta\right)
    \right]
\end{align}
where $\mathcal{I}(Q)$ is an indicator function that equals to $1$ if $Q$ is true and $0$ otherwise. We note that directly optimizing $\text{SR}@\eta$ is difficult due to its non-differential property. To address this issue,
in the following section, we propose a decomposition-based model.
Moreover, in order to enhance robustness against extreme events, we introduce a self-adaptive loss that can adaptively assign weights according to the fitting error of each time stamp.

In Figure~\ref{fig:success_rate}, we plot the success rate defined in Equation~\ref{equ:success_rate} with various thresholds $\eta \in [0.8,1]$. It is observed that the success rate of SaDI outperforms all other baselines with a range of different $\eta$.

\subsection{Ablation Study}
\input{tables/tab_ablation.tex}
Table~\ref{tab:ablation_study} shows that the ablation of any module will degrade the performance. It is worth noting that we implement GAM using GBDT with a depth of 1. GAM has worse performance than deeper boosting tree model. Thus, SaDI with GAM sacrifices the performance to gain interpretability.

\subsection{Evaluation of Speed}
\label{app:speed}
\input{tables/tab_bench_evtloss2}
As shown in Table \ref{tab:benchmark_speed}, SaDI has the shortest training speed among all baselines. The integrated GAM in SaDI is light-weighted. SaDI has a higher inference time since the three decomposed issues will be predicted individually and then summed up. However, the long inference time is acceptable.

%% file: sections/1.5_related_work.tex
\section{Related Work}
\label{app:related_work}

\subsection{Time Series Methods}
Numerous research works have been involved in load forecasting in recent years. For cases without extreme events, the forecasting technology both in research~\cite{nbeats/iclr2019} and industry~\cite{tcn_electric/2021} is mature. The traditional machine learning methods have limitations in representation power. While, Neural Networks (NN), although being widely investigated in time series tasks, are time-consuming and not suitable for mass deployment. We summarize the challenges and limitations of general time series models in Table \ref{tab:challenge}.

\input{tables/tab_challenge}


\subsection{Statistic Methods}
Some recent studies in statistics have shown their superiority in dealing with extreme events. Exponential smoothing
~\cite{Holt-winters/STLP/} is widely used in forecasting since it is capable of capturing trend and seasonal characteristics. However, its performance degrades when the forecast horizon increases or some change points information is not perceived. 
Recently, Extreme Value Loss (EVL)~\cite{EVLoss/kdd19/} uses Extreme Value Theory (EVT) to detect the possible future occurrences of extreme events.
However, EVL is designed based on the assumption that samples are independent and identically distributed (i.i.d), which is seldom satisfied for time series data. Empirical results have shown that the EVL-based method works ordinarily. 


\subsection{Imbalanced Regression Methods}
Work of~\cite{General_Imbalanced_regression/IR/} presents a new approach to dealing with extreme events from the perspective of imbalanced regression~(IR), where the objective is to predict extreme values via relevance functions with more attention paid to extreme events through reweighting. 
Deep Imbalanced Regression combines IR and deep learning to learn continuous targets from naturally imbalanced data. 
In this model~\cite{IR_LDS/ICML21/}, kernel methods are used to smooth label, and feature distributions are learned. 
However, the skew in distributions of label and features are solved independently, where no explicit relationship is made between the features and the label under extreme events. 

%% file: tables/tab_challenge.tex
\begin{table}[h]
\centering
\begin{footnotesize}
\caption{Analysis of challenges. LR: linear regression models. Tree: tree-based boosting models. NN: neural networks. SaDI: our method. ``+'' means the model is capable of handling the challenge. ``-'' means incapable. ``++'' means the capacity is prominent.}
\scalebox{0.9}{
\begin{tabular}{c|cccc}
\hline
Challenge & LR & Tree & NN & SaDI  \\ 
\hline
Interpretable &  ++ & + & - & ++ \\

Extrapolation & + & - & - & + \\

Fast-adaptation & - & - & - & + \\

Representation power & + & ++ & ++ & ++ \\

Require CPU only &   ++ & + & - & +\\ 
\hline
\end{tabular}
\label{tab:challenge}
}
\end{footnotesize}
\end{table}

%% file: tables/tab_dataset.tex
\begin{table}[h]
\centering
\begin{footnotesize}
\caption{Summary of datasets.}
\scalebox{0.9}{
\begin{tabular}{c|c|cccc}
\hline
\multicolumn{2}{c|}{Dataset}& \#Features & \#Samples & Sample rate \\ 
\hline
\multirow{4}{*}{Huazhong} 
& Hubei & 14 & 86113 & 15 min \\ 
         & Hunan & 14 & 86113 & 15 min \\ 
         & Henan & 14 & 86113 & 15 min \\ 
         & Jiangxi & 14 & 86113& 15 min \\ 
\hline
         


Public & Peacock & 9 & 70173 & 15 min \\ 
& Rat& 9 & 70173 & 15 min \\ 
& Robin & 9 & 70173 & 15 min \\ 
\hline

         

\end{tabular}
\label{tab:dataset}
}
\end{footnotesize}
\end{table}


%% file: tables/algorithm1.tex

\RestyleAlgo{ruled}
\SetKwInOut{Input}{Input}\SetKwInOut{Output}{Output}
\SetKw{Define}{Define}
\SetKw{Return}{Return}
\SetKw{Initialize}{Initialize}
\SetKwFunction{SaDI}{SaDI}

\begin{algorithm}[h]
\fontsize{10}{12}\selectfont
\caption{SaDI}\label{alg:selectfeatures}
\Input{$\{\boldsymbol{x}_t,y_t\}_{t=0}^{N-1}$: A set of training samples.\\ $\{\boldsymbol{x}_t\}_{t=N}^{N+m-1}$: features to predict.}
\Output{Prediction $\{y_t\}_{t=N}^{N+m-1}$.}
\tcc{Performing decomposition}
$y_t^{LT}\leftarrow \text{moving average of}\; y_t$\\
$y_t^{ST}\leftarrow \text{moving average of}\; y_t-y_t^{LT}$\\
$y_t^{S}\leftarrow y_t-y_t^{LT}-y_t^{ST}$\\
\tcc{Modeling $y_t^{LT}$ linear model}
$\boldsymbol{w}^*\leftarrow \arg\min \sum_{t=0}^N (y_t^{LT}-\boldsymbol{w}^T\boldsymbol{x}_t)^2$\\
\tcc{Learn $y_t^{ST}$ with GAM}
$\text{Model1}\leftarrow \text{GAM( training set =}$ $\{\boldsymbol{x}_t,y_t^{ST}\}$, $\text{loss=ETL)}$\\
\tcc{Modeling $y_t^S$ with lightGBM}
$\text{Model2}\leftarrow \text{lightgbm( training set =}$ $\{\boldsymbol{x}_t,y_t^{S}\}$, $\text{loss=RMSE)}$\\
\tcc{Predict using learnt models}
$\hat{y}_t^{LT}\leftarrow (\boldsymbol{w}^*)^T\boldsymbol{x}_t\quad \forall t=N\dots N+m-1$\\
$\hat{y}_t^{ST}\leftarrow \text{Model1} (\boldsymbol{x}_t)\quad \forall t=N\dots N+m-1$\\
$\hat{y}_t^{S}\leftarrow \text{Model2} (\boldsymbol{x}_t)\quad \forall t=N\dots N+m-1$\\
$y_t \leftarrow \hat{y}_t^{LT}+\hat{y}_t^{ST}+\hat{y}_t^{S}\quad \forall t=N\dots N+m-1$\\
\Return{$\{y_t\}_{t=N}^{N+m-1}$}
\end{algorithm}


%% file: tables/main_procedures.tex

\RestyleAlgo{ruled}
\SetKwInOut{Input}{Input}\SetKwInOut{Output}{Output}
\SetKw{Define}{Define}
\SetKw{Return}{Return}
\SetKw{Initialize}{Initialize}
\SetKwFunction{SaDI}{SaDI}

\begin{algorithm}[h]
\centering
\fontsize{10}{12}\selectfont
\caption{Framework Experiments Procedure}\label{alg:selectfeatures}

\text{DateTimeFeaturizer}\\
\text{DifferenceFeaturizer}\\
\text{RollingStatsFeaturizer}\\
\text{FeatureEnsembler}\\
\text{Pred\_one\_component}\\

\end{algorithm}
The algorithm mentioned is according to "SaDI\_demo.ipynb" codes enable to help understand our "SaDI" framework.

$DateTimeFeaturizer$ is used for constructing temporal features with timestamps,such as year, month, day. And some extra-features, such as is workday or not, also can derive from calendar's information.

$DifferenceFeaturizer$ is used for loading difference weather features ,such as "2 meter temperature", "Surface pressure" numerical weather prediction (NWP) attributes. The incremental of these attributes is related to the change of the load. Hence, we do the difference operator to the NWP attributes 2 days before(always 192 points ahead).

$RollingStatsFeaturizer$ is used for rolling history load data with different windows, such 1 or 7. Rolling history load represents the past loads influences to the future. It is essential for algorithm to learn historical patterns of load.

$FeatureEnsembler$ is used for ensembling temporal, weather and load features together and then over-all features being inputs can drive model to predict future load.

$Pred\_one\_component$ is used for predicting long-term trend, short-term trend,period components with different parameters. When to forecast long-term trend the parameter $model='linear'$. In the case of predicting short-term trend the parameter $model='evloss'$. For forecasting period, the parameter $model='lgb'$.
The final predicted value is the sum of above components.


%% file: tables/tab_ablation.tex
\begin{table}[t]
\centering
\begin{footnotesize}
\caption{Ablation study of SaDI on pulic dataset.}
\scalebox{0.7}{
\begin{tabular}{c|cc|cc|cc}
\hline
\multicolumn{1}{c|}{Datasets}&\multicolumn{2}{c|}{Peacock}&\multicolumn{2}{c|}{Rat}&\multicolumn{2}{c}{Robin}\\
\hline
\hline
\multicolumn{1}{c|}{Metrics} & $\rm{nRMSE_d}$ & $\rm{MAPE_d}$ & $\rm{nRMSE_d}$  & $\rm{MAPE_d}$ & $\rm{nRMSE_d}$ & $\rm{MAPE_d}$\\
\hline
SaDI & 0.0250 & 0.0208 & 0.0855 & 0.0772 & 0.0682 & 0.0599\\


SaDI wo deompose & 0.0579 & 0.0464 & 0.1628 & 0.1515 & 0.1151 & 0.1014 \\


SaDI wo Feature engineer & 0.2192 & 0.2088 & 0.3181 & 0.2900 & 0.1566 & 0.1353 \\

SaDI wo EVL & 0.0254 & 0.0212 & 0.0914 & 0.0835 & 0.0691 & 0.0609 \\

SaDI w GAM & 0.0284 & 0.0234 & 0.0901 & 0.0798 & 0.0737 & 0.0642\\

\hline
\end{tabular}
\label{tab:ablation_study}
}
\end{footnotesize}
\end{table}

%% file: tables/tab_bench_evtloss2.tex
\begin{table}[h]
\centering
\begin{footnotesize}
\caption{Comparison of SaDI and baselines on training speed and inference speed.} 
\begin{tabular}{c|c|c|c|c}
\hline
Methods & SaDI & LightGBM & EVT & LDS \\
\hline
\hline
Training time (s) & \textbf{71.8}  & 88.7  & 225.7 & 110.2  \\

Inference time (s) & 0.972  & 0.695  &\textbf{0.07} & 1.51  \\
\hline












\end{tabular}
\label{tab:benchmark_speed}
\end{footnotesize}
\end{table}